\pdfoutput=1

\documentclass[11pt]{article}

\usepackage[final]{ACL2023}

\usepackage{times} \usepackage{latexsym} \usepackage{amsmath} \usepackage{tikz}
\usepackage{amssymb} \usepackage{hyperref} \usepackage{graphicx}
\usepackage{titlesec}
\usepackage{hyperref}
\usepackage{cprotect}
\usepackage{float}
\usepackage[utf8]{inputenc}
\usepackage[T1]{fontenc}
\usepackage{expex}

\titleformat*{\section}{\Large\bfseries}
\titleformat*{\subsection}{\large\bfseries}
\titleformat*{\paragraph}{\normalsize\bfseries}

\usepackage[T1]{fontenc}

\usepackage[utf8]{inputenc}

\usepackage{microtype}

\usepackage{inconsolata}

\renewcommand{\vec}[1]{\mathbf{#1}}
\title{CLaC at SemEval-2023 Task 2: Comparing Span-Prediction and
Sequence-Labeling approaches for NER}

\author{Harsh Verma, Sabine Bergler  \\
        CLaC Labs, Concordia University \\
        \{h\_ver, bergler\} @cse.concordia.ca}

\begin{document} 
\maketitle 
\begin{abstract} This paper summarizes the CLaC
submission for the MultiCoNER 2 task which concerns the
recognition of complex, fine-grained named entities. 
We compare two popular approaches for NER, namely Sequence
Labeling and Span Prediction. We find that our 
best Span Prediction system performs slightly
better than our best  Sequence Labeling system on test data.
Moreover, we find that using the larger version of XLM RoBERTa significantly improves performance. Post-competition experiments show that Span Prediction and Sequence Labeling approaches improve 
when they use special input tokens (\verb|<s>| and \verb|</s>|) of XLM-RoBERTa. The code for training all models, preprocessing, and post-processing is available at \href{https://github.com/harshshredding/semeval2023-multiconer-paper}{this Github repo}.
\end{abstract}

\section{Introduction} \label{intro} MultiCoNER 2 \cite{multiconer2-report,multiconer2-data} is
the second version of the MultiCoNER \cite{multiconer-report} shared task.
It focuses on identifying complex named entities in short
sentences that lack context and addresses the challenge of identifying and distinguishing 30 fine-grained entity types and of handling simulated errors, such as typos. 

Competition data is tokenzied and tagged with the BIO scheme, i.e. the label B for \emph{begin}, I for \emph{inside}, and O for \emph{outside} target entities. For this challenge, the 30 entity types were distinguished using suffixes to the B and I labels, e.g. B-Corp and I-Corp, as illustrated in Example~\ref{bio-ex}.

NER is frequently formulated as
a sequence labeling problem \cite{chiu-nichols-2016-named, ma-hovy-2016-end, wang-etal-2022-damo}, in which a model 
learns to label each token individually using  the BIO labeling scheme. In contrast, span prediction approaches 
\cite{jiang-etal-2020-generalizing, li-etal-2020-unified, spanner, zhang2023optimizing} label entire text spans with entity types. 
Recently, \cite{spanner} compared the two approaches and found 
that sequence labeling performed better for long  entities with low label consistency but that span prediction performed better for out of vocabulary words and entities of short to medium length.
Because MultiCoNER 2 tests both, the ability to identify out of vocabulary entities and entities of varying length\footnote{Creative work titles can be long, e.g. \emph{To Kill A Mocking Bird}}, it is interesting to compare the two approaches on this  task.

 Our submission to the English Track does not use any external knowledge and holds rank 21 of 41 on Codalab\footnotemark{} with a macro-F1 of 59. \footnotetext{\url{https://codalab.lisn.upsaclay.fr/competitions/10025\#results}}

\section{Submitted Systems} 
\subsection{Tokenization} The competition input consisted of  tokenized data. The gold labels are given in form of BIO tags, 
where each token is labeled with a corresponding BIO tag as shown in Example~\ref{bio-ex}.
Sample input is given to  XLM RoBERTa \cite{xlm} 
as the raw input string, without RoBERTa input tags such as
\verb|<s>| and \verb|</s>| in the submission systems.




\ex  \label{bio-ex}
\begingl 
\gla \textbf{Token} pharma company eli lilly and company announced //
\glb \textbf{BIO-Tag} O O \texttt{B-Corp} \texttt{I-Corp} \texttt{I-Corp} \texttt{I-Corp} O  //
\endgl
\xe

\subsection{Sequence Labeling Model} 

Our sequence labeling model uses  XLM RoBERTa large \cite{xlm}. Like all BERT-derived models, RoBERTa splits the input tokens further into word piece subtokens, for which the pre-trained embedding vectors are included, and for which the subtoken-level output is generated. These subtokens are classified into the BIO sequence. For submission and scoring, this subtoken sequence has to be decoded into word-level tokens.

\paragraph{Token Representation Step} Given a sentence $\vec{x} = [w_{1},
w_{2}, ..., w_{n}]$ with $n$ word piece tokens, we generate for each token $w_{i}$ 
a pretrained embedding $\vec{u}_{i}$ using the 
last hidden layer of XLM RoBERTa large \cite{xlm} 

\begin{align*} 
\text{Embed}(\textbf{x}) &= \text{Embed}([w_{1}, w_{2}, ..., w_{n}]) \\
                            &= [\vec{u_{1}}, \vec{u_{2}}, ..., \vec{u_{n}}] \\
\end{align*}

\paragraph{Token Classification Step} In this layer, we classify every token
representation into a set of named entity types corresponding to the
BIO(\textit{beginning}, \textit{inside}, \textit{outside}) tagging scheme.
Assuming $\mathbf{\Theta}$ is the set of all named entity types 
, the set of all BIO tags $\mathbf{B}$ is of size $(2 \times |\vec{\Theta}|) + 1$. We use a linear layer to map each subtoken $\vec{u}_{i} \in
\mathbb{R}^{d}$ to a prediction $\vec{p}_{i} \in \mathbb{R}^{|\bf{B}|}$, where
$d$ is the length of the token embedding. The predictions are used to
calculate loss of sample $\vec{x}$ with $n$ tokens as follows:
\begin{equation} \text{Loss}(\vec{x}) =
\frac{-1}{n}\sum_{i=1}^{n}\text{log}(\text{Softmax}(\vec{p}_{i})_{y_{i}})
\end{equation} Here $y_{i}$ represents the index of the gold BIO label of the
$i^{th}$ token.

\paragraph{Decoding Step} For this task, the boundaries of each predicted span must
align with actual word boundaries, which poses an issue due to the word piece tokenization. 
We align every predicted span with the nearest enveloping words. Concretely, let $(b, e)$ represent the beginning and end offsets of a predicted span $s$. If $b$ and $e$ are contained in words $w_{b}$
and $w_{e}$ respectively, they are remapped to the beginning and end offsets of the containing words. 

\subsection{Span Prediction Model}

\paragraph{Token Representation Layer} The token representation layer is identical to that of the Sequence
Labeling model.

\paragraph{Span Representation Layer} 
Let a span $\vec{s}$ be a tuple $\vec{s} = (b,e)$ where $b$ and $e$ are the
begin and end word piece token indices, and $\vec{s}$ represents the text segment
$[w_{b}, w_{b+1}, ..., w_{e}]$ where $w_{i}$ is the $i^{th}$ word piece token.
In this layer, we enumerate all possible spans, represented by the tuple $(b,e)$. 
Because $b \leq e$, there are $\frac{n^{2}}{2}$
possible spans.
We follow \cite{spanner} and encode each span $s_{i}$ as the concatenation of their begin and end word piece token embeddings $\vec{v}_{i} = [\vec{u}_{b_{i}};\vec{u}_{e_{i}}]$. The output of the decoding layer is
$\textbf{V} \in \mathbb{R}^{k \times (2 \times d)}$ where $k = \frac{n^{2}}{2}$
and $d$ is length of the token embedding vector.

\paragraph{Span Classification Layer} In this layer, we classify each span representation   with a named entity type 
. We introduce an additional label
\verb|Neg_Span| which represents the absence of a named entity 
. In particular, a
linear layer maps each span representation $\vec{v}_{i} \in \mathbb{R}^{(2
\times d)}$ to a prediction $\vec{p}_{i} \in \mathbb{R}^{|\Omega|}$
, where
$\Omega$ is the set of all named entity types (including \verb|Neg_Span|) and
$d$ is the size of the token embedding. The predictions are used to calculate the loss for sentence $\vec{x}$ with $l$ possible spans as follows:
\begin{equation} \text{Loss}(\vec{x}) =
\frac{-1}{l}\sum_{i=1}^{l}\text{log}(\text{Softmax}(\vec{p}_{i})_{y_{i}})
\end{equation} Here $y_{i}$ represents the index of the gold label of the
$i^{th}$ span.

\paragraph{Decoding} Similarly to the Sequence Labeling model, we align all spans with word boundaries. Because the 
Span Prediction model predicts overlapping spans, we  remove overlaps with the following procedure:
\begin{enumerate}
    \item For each span $s$, if $s$ is completely contained within another span $S$, we remove $s$. We keep removing fully contained spans until none are left.
    \item For each span $s_{1}$ which partially overlaps  another span  $s_{2}$, we randomly select one of the two spans and remove it. We keep
removing until no overlapping spans are left. 
\end{enumerate}

The remaining spans are then mapped to BIO tags
.

\subsection{Training} XLM RoBERTa large is fine-tuned on the training data using the Adam optimizer \cite{adam} with a learning rate of \verb|1e-5| and a batch size of 1. The best model is selected using early stopping. Training for 10 epochs takes around 6 hours on one Nvidia RTX 3090 gpu.

\section{Results and Discussion} 

We evaluate all of our systems on the Codalab Competition Website\footnote{\url{https://codalab.lisn.upsaclay.fr/competitions/10025}}, where submissions are evaluated using the entity level macro F1 metric. 
All models use the XLM-RoBERTa large LM unless otherwise indicated.

Table~\ref{table:comprehensive}  shows the performance of the two submitted systems S(submitted) in bold.

Post-competition experimentation showed that increasing the batch size from 1 to 4 improved results   by at least $3.2$, this improvement is shown in Table~\ref{table:comprehensive} for systems B.

We also found that  adding the embeddings for RoBERTa special tokens \verb|<s>| and \verb|</s>| further  improved performance, especially raising performance of  the Sequence model Seq  by $2.1$, nearly erasing the performance advantage of the span-based model. The results are shown in Table~\ref{table:comprehensive} for systems B+E.


\begin{table}[htb]
\scalebox{0.87}{ 
\begin{tabular}{l|lll|lll}
System Type & \multicolumn{3}{c|}{Seq} & \multicolumn{3}{c}{Span} \\ \hline
 & \multicolumn{1}{l|}{F1} & \multicolumn{1}{l|}{P} & R & \multicolumn{1}{l|}{F1} & \multicolumn{1}{l|}{P} & R \\ \hline
S(submitted) & \multicolumn{1}{l|}{\bf 53.2} & \multicolumn{1}{l|}{\bf 52.8} & {\bf 54.5} & \multicolumn{1}{l|}{\bf 55.0} & \multicolumn{1}{l|}{\bf 56.5} & {\bf 55.5} \\ \hline
B & \multicolumn{1}{l|}{56.9} & \multicolumn{1}{l|}{57.3} & 57.5 & \multicolumn{1}{l|}{59.0} & \multicolumn{1}{l|}{61.2} & 58.4 \\ 
B+E & \multicolumn{1}{l|}{59.0} & \multicolumn{1}{l|}{60.7} & 58.9 & \multicolumn{1}{l|}{59.7} & \multicolumn{1}{l|}{61.3} & 59.5 \\ 
\end{tabular}
}
\caption{
    Performance on the test set in bold. The submitted systems improve when the batch
    size is increased from 1 to 4 (systems B), and they improve further when 
    special RoBERTa token embeddings are  added (systems B+E).
} 
\label{table:comprehensive} 
\end{table}

Table~\ref{table:perf_validation} shows the performance of 
our systems on the validation set.

The Span Prediction models showed significantly higher performance on the validation set than the test set. The post-competition improvements affect the sequence and span-based models differently and nearly erase the performance difference.

\begin{table}[htb] \scalebox{0.87}{ 
\begin{tabular}{|p{2.4cm}|l|l|l|l|} \hline
    Model            & $\text{Span}_{\text{S}}$  & $\text{Seq}_{\text{S}}$ & $\text{Span}_{\text{B}}$  & $\text{Seq}_{\text{B}}$ 
    \\ \hline Macro F1 & 58.5 & 59.3 & 63.6 & 60.2
    \\ \hline  

\end{tabular} } \caption{Performance on validation set for systems S and B} 
\label{table:perf_validation}
\end{table}

Table~\ref{table:compare_large_small} compares the
performance of large and base versions of XLM RoBERTa. $\text{Span}_{\text{base}}$ and
$\text{Seq}_{\text{base}}$ are identical to
$\text{Span}_{\text{B}}$ and  $\text{Seq}_{\text{B}}$ 
except that the base models use XLM RoBERTa \textit{base} instead of large.
The larger models seem to be performing significantly better. 


\begin{table}[htb] \scalebox{0.87}{ 
\begin{tabular}{|p{2.4cm}|l|l|l|l|} \hline
        Model            & $\text{Span}_{\text{B}}$  &
$\text{Seq}_{\text{B}}$  & $\text{Span}_{\text{base}}$ &
$\text{Seq}_{\text{base}}$  \\ \hline Macro F1 & 59.0 & 56.9 & 52.5 & 51.1
    \\ \hline  

\end{tabular} } 
\caption{Performance of large and base
pretrained models on test set for systems B  } 
\label{table:compare_large_small} 
\end{table}




\section{Conclusion} 

We submitted two systems to MultiCoNER 2, 
one inspired by a Sequence Labeling approach and another inspired by a Span Prediction approach. We find that our best Span Prediction system performs slightly
better than our best Sequence Labeling system on test data. 
We showed significant increases in our systems' performance when using a larger pretrained language model, batch size 4, and special tokens \verb|<s>| and \verb|</s>|.


\bibliography{anthology,custom} \bibliographystyle{acl_natbib}



\end{document}